\documentclass{article}

\usepackage[preprint,nonatbib]{neurips_2024}

\usepackage[utf8]{inputenc} 
\usepackage[T1]{fontenc}    
\usepackage{hyperref}       
\usepackage{url}            
\usepackage{booktabs}       
\usepackage{amsfonts}       
\usepackage{nicefrac}       
\usepackage{microtype}      
\usepackage{xcolor}         
\usepackage{array}
\usepackage{graphicx}
\usepackage{multirow}
\usepackage{subcaption}
\usepackage{algorithm}
\usepackage{algpseudocode}
\usepackage{amsmath,amssymb}
\usepackage{wrapfig}

\title{Rethinking the Potential of Layer Freezing for Efficient DNN
Training}

\author{Chence Yang\\
University of Georgia\\
\texttt{cy65664@uga.edu} \\
\And
Ci Zhang\\
University of Georgia\\
\texttt{cz06540@uga.edu} \\
\And
Lei Lu\\
Northeastern University\\
\texttt{lu.lei1@northeastern.edu} \\
\And
Qitao Tan\\
University of Georgia\\
\texttt{email} \\
\And
Sheng Li\\
University of Pittsburgh\\
\texttt{shl188@pitt.edu} \\
\And
Ao Li\\
The University of Arizona\\
\texttt{aoli1@arizona.edu} \\
\And
Xulong Tang\\
University of Pittsburgh\\
\texttt{xulongtang@pitt.edu} \\
\And
Shaoyi Huang\\
Stevens Institute of Technology\\
\texttt{shuang59@stevens.edu} \\
\And
Jinzhen Wang\\
The University of North Carolina at Charlotte\\
\texttt{jwang96@charlotte.edu}\\
\And
Guoming Li\\
University of Georgia\\
\texttt{gmli@uga.edu} \\
\And
Jundong Li\\
University of Virginia\\
\texttt{jundong@virginia.edu} \\
\And
Xiaoming Zhai\\
University of Georgia\\
\texttt{xiaoming.zhai@uga.edu} \\
\And
Jin Lu\\
University of Georgia\\
\texttt{jin.lu@uga.edu} \\
\And
Geng Yuan\\
University of Georgia\\
\texttt{geng.yuan@uga.edu} \\
}

\begin{document}

\maketitle

\begin{abstract}
With the growing size of deep neural networks and datasets, the computational costs of training have significantly increased. 
The layer-freezing technique has recently attracted great attention as a promising method to effectively reduce the cost of network training.
However, in traditional layer-freezing methods, frozen layers are still required for forward propagation to generate feature maps for unfrozen layers, limiting the reduction of computation costs. To overcome this, prior works proposed a hypothetical solution, which caches feature maps from frozen layers as a new dataset, allowing later layers to train directly on stored feature maps. While this approach appears to be straightforward, it presents several major challenges that are severely overlooked by prior literature, such as how to effectively apply augmentations to feature maps and the substantial storage overhead introduced.
If these overlooked challenges are not addressed, the performance of the caching method will be severely impacted and even make it infeasible.
This paper is the first to comprehensively explore these challenges and provides a systematic solution. To improve training accuracy, we propose \textit{similarity-aware channel augmentation}, which caches channels with high augmentation sensitivity with a minimum additional storage cost. 
To mitigate storage overhead, we incorporate lossy data compression into layer freezing and design a \textit{progressive compression} strategy, which increases compression rates as more layers are frozen, effectively reducing storage costs. Finally, our solution achieves significant reductions in training cost while maintaining model accuracy, with a minor time overhead. Additionally, we conduct a comprehensive evaluation of freezing and compression strategies, providing insights into optimizing their application for efficient DNN training. Our code is released at \url{https://anonymous.4open.science/r/Freezing_and_discarding-14C3}.
\end{abstract}

\section{Introduction}
In recent years, with the ever-increasing size of deep neural networks (DNNs) and datasets, the training costs of DNNs have risen significantly. Since early layers in DNNs typically have fewer parameters and focus on extracting low-level features, they tend to converge faster than the later layers during the training process \cite{b15}. Therefore, layer-freezing techniques have emerged as a promising approach to reducing the computational costs of DNN training. Specifically, the layer freezing technique stops the weight updates for certain layers during training, which can be considered as ``freezing'' those layers. As a result, frozen layers no longer require backpropagation to compute gradients for weight updates, thereby reducing computational overhead.

However, existing layer-freezing techniques still have limitations. One of them is that even frozen layers (i.e., earlier layers) still need to perform computations for forward propagation. This is because the input to the unfrozen layers (i.e., later layers) depends on the output (i.e., activations/feature maps) of the frozen layers, which requires forward propagation through the frozen layers to obtain. As a result, the forward computation cost of frozen layers cannot be saved, leaving a particularly unfortunate imperfection in layer freezing techniques.

Intriguingly, considering that the weights of frozen layers remain unchanged, during the iterative training process, when the same training data (or data from the same dataset) is fed into the frozen layers, the output feature maps generated by the frozen layers should ideally remain the same. This provides the possibility for layer-freezing techniques to further eliminate the need for forward propagation.
Therefore, some of the previous works  \cite{b8,b12} proposed a hypothetical solution: \textit{caching the outputs.} Specifically, the caching method stores the outputs (e.g., activations/feature maps) of the (last) frozen layer as a new dataset on disk storage. During subsequent training, the frozen layers can be discarded, while the unfrozen layers directly load the cached outputs and use them as a new dataset to continue training. If this method is feasible, it would enable layer-freezing techniques to completely eliminate the need for both forward and backward propagation calculations for frozen layers, thereby fully realizing the potential computational savings.
Previous works on layer freezing primarily focused on designing freezing criteria and algorithms. The caching method was only briefly discussed as a hypothetical approach, making it appear to be a very simple and effective solution. However, \textit{is the caching method truly as simple and effective in practical applications as it seems?}

In fact, our research reveals that the caching method presents several essential challenges that, if not effectively addressed, can severely impact its feasibility and potential. 
The first challenge is how to effectively perform data augmentation on the cached feature maps to ensure the network's performance. 
Note that, due to the nonlinear nature of neural networks and the fact that the output of later layers typically represents high-level feature maps, traditional data augmentation techniques that work well on the original input data may not directly apply to the feature maps from later layers. This inability to effectively augment the new cached dataset could lead to a severe decrease in the model accuracy.
Another critical challenge is that
activations/feature maps typically have a larger size than the original input data due to the increased dimensionality and higher representation precision (e.g., from integers to floating-point numbers). If stored in their entirety as a new dataset, this would result in substantial storage requirements, making the method impractical for real-world implementations.
For example, a 224$\times$224$\times$3 image passes through the first residual block in ResNet50 will become a tensor with a dimension of 56$\times$56$\times$256, and the type changes from Int8 to Float32. This makes the newly stored dataset 21.3$\times$ larger than the original dataset.

In this work, we conducted a comprehensive exploration to address the aforementioned challenges. 
First, we found that after applying the caching method in layer-freezing and discarding frozen layers, directly performing augmentations similar to those applied to the original input (e.g., crop, flip, color jittering) on the loaded feature maps leads to significant accuracy degradation in the final trained model. Therefore, we propose a novel \textit{similarity-aware channel augmentation} method that effectively enhances the accuracy of model training. 
Then, to address the issue of the excessive dataset size for storing feature maps, we propose incorporating lossy data compression in layer-freezing. We further explore its feasibility and its impact on the accuracy of model training. We found that the feature extraction characteristics of neural networks make data compression algorithms highly compatible with the caching method in layer-freezing. We propose a progressive compression method, which gradually increases the compression rate as the number of frozen layers increases. This approach minimizes the storage requirements for the newly generated dataset while maintaining the accuracy of model training.
Moreover, we further explore the impact of data compression quality on the model accuracy, memory cost, and training speed for different scenarios.

Based on our comprehensive study, we have drawn several interesting and insightful conclusions. 
Though the caching technique for layer freezing will slightly compromise the training speed due to the limited computing speed of current data compression techniques. And well-designed feature augmentation methods are required to maintain model accuracy.
Even so, it is still a promising solution that can significantly save memory costs and training FLOPs for both CNN-based and transformer-based models, which is particularly valuable for efficiency considerations.
Moreover, with the advancement of data compression techniques in the future, the potential of the caching technique for layer freezing will be further unleashed.

The key contributions of this paper are summarized as follows:

\begin{itemize}
    \item We are the first study to comprehensively explore the difficulties and potentials of implementing caching techniques for layer-freezing.
    \item We propose a similarity-aware channel augmentation method to enhance the effectiveness of training on cached feature maps.
    \item We propose a progressive compression strategy, which reduces storage and computational overhead while maintaining model performance.
    \item We analyze compression rates and freezing strategies across different depths and training stages to optimize efficiency and accuracy.
    \item Our proposed methods make the caching technique a feasible solution for layer-freezing. Our results show that our method can save up to \textbf{24.4}\% training FLOPs and \textbf{48.4}\% memory usage without introducing considerable model accuracy degradation and training time overhead. 

\end{itemize}

\section{Background}
\subsection{Layer Freezing}
Some research has shown that not all layers in a deep neural network require equal training effort \cite{b13}. For example, in CNN-based models, lower layers typically capture low-level features such as textures, corners, and edges \cite{b4}. These layers, having fewer parameters, can quickly learn fundamental representations and often do not require continuous updates throughout training. Building on this observation, layer freezing has been proposed as an effective technique to optimize training efficiency by selectively halting parameter updates in specific layers while keeping others trainable. This approach is particularly beneficial in transfer learning, where pretrained lower layers already encode general features \cite{b14,b20}.

To improve efficiency, several prior studies focus on developing better freezing criteria.
Egeria \cite{b8} dynamically determines when to freeze layers based on their training progress, reducing unnecessary computations. Similarly, SmartFRZ \cite{b9} employs attention mechanisms to identify which layers can be frozen, ensuring that training remains efficient without sacrificing performance. Moreover, SpFDE \cite{b10} explores the compatibility of layer-freezing with sparse training techniques. 
Prior work \cite{b8,b12} proposes caching the outputs of the frozen layers as a hypothetical solution to eliminate the forward propagation in the frozen layers.
However, a severe accuracy degradation can be observed \cite{b8}, and it only works when using a very small dataset \cite{b12}. These issues raise concerns about the feasibility and potential of the caching method.

\subsection{Feature Map Augmentation}
Data augmentation is a widely used technique in DNN training. Common methods include geometric transformations such as random cropping, flipping, rotation, and color jitter \cite{b22,b23}. These transformations help models maintain robust performance across variations in perspective, scale, and position, preventing overfitting and promoting the learning of generalizable features.
However, in specific scenarios such as medical image analysis, traditional augmentation methods may alter inherent spatial relationships, making direct geometric transformations unsuitable. Some studies explore applying spatial transformations and noise perturbations directly in feature space to achieve single-source domain generalization and improve cross-domain robustness \cite{b2,b3,b26}. Some works also propose augmentations in hidden layers for better generalization \cite{b25}.
Nonetheless, our study found that directly applying existing augmentation methods to cached data (e.g., feature maps) in layer-freezing still fails to prevent accuracy degradation.

\begin{figure*}[!ht]
\centering  
\includegraphics[width=\textwidth]{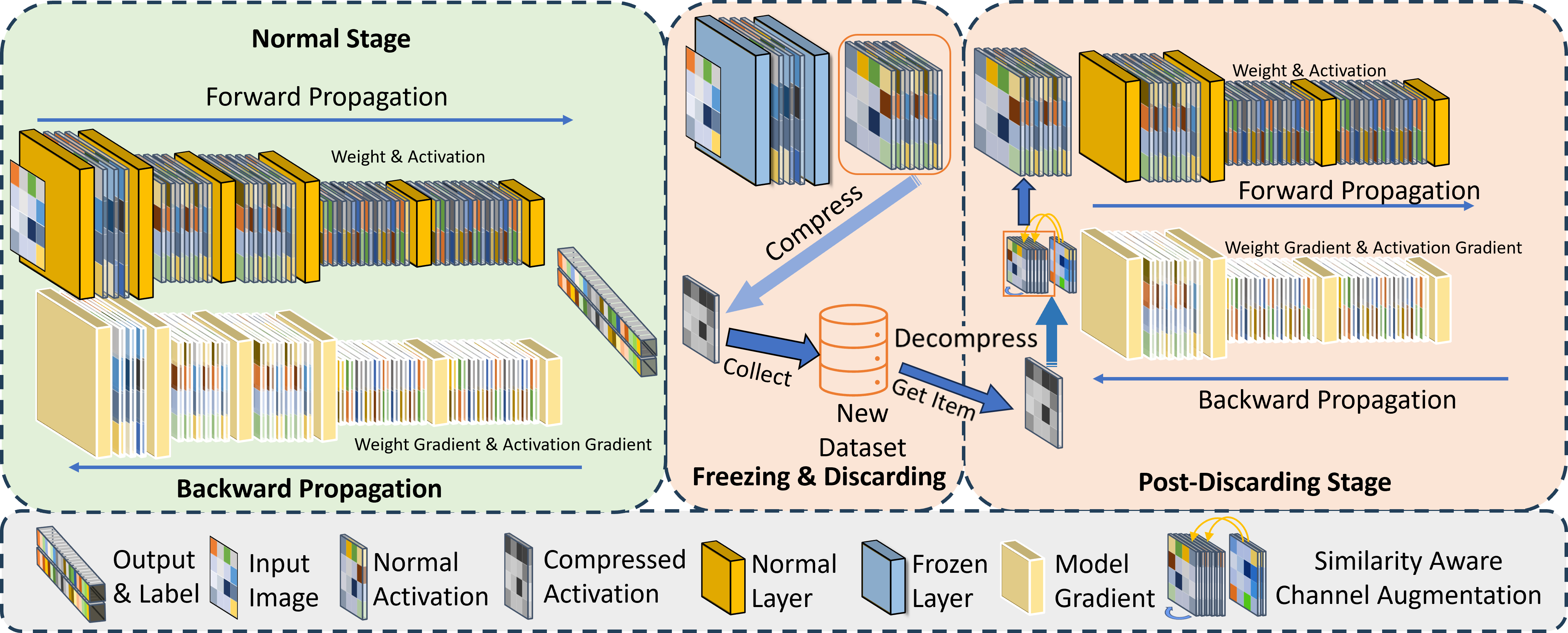}
\caption{The overview of how to discard the frozen partial of model and applying the cached feature map augmentation. 
At the designated freezing point, the model is divided into a frozen part and an unfrozen part. All training data is passed through the frozen part to generate activations, which are then compressed and cached, and the frozen section subsequently discarded. During the subsequent training of the unfrozen part, the required training data is decompressed and subjected to relevant feature map augmentation.
}

\label{fig:pipeline}
\end{figure*}
\subsection{Data Compression}
With the exponential growth of digital data, lossy data compression has been widely used for optimizing storage and transmission efficiency. 
Lossy compression generally achieves a higher compression rate than lossless compression by discarding non-essential information.
However, traditional lossy techniques (e.g., DCT and DWT) are less effective for scientific data, which is high-precision, floating-point, large-scale, and high-dimensional. These methods struggle to maintain precision and often introduce non-linear errors, making them unsuitable for numerical simulations and scientific analysis. To address this, ZFP \cite{b21} has emerged as a specialized compression method, utilizing block-based transform coding to achieve high compression while ensuring controllable error bounds. To further enhance precision, NeruLZ \cite{b17} employs a learnable model to predict and correct compression errors. Additionally, some studies have explored compressing activations during training and decompressing them for gradient computation, effectively reducing memory usage \cite{b5,b6,b7}. Given the lower compression ratios and slower processing speeds of lossless methods—along with their significant memory and time overhead—and the fact that intermediate data after lossy compression can still retain adequate accuracy, we adopt ZFP \cite{b21} as a proof-of-concept to demonstrate the effectiveness and compatibility of layer freezing with lossy compression.

\vspace{-1em}

\section{Methodology}

\subsection{Revisit Frozen Layer Discarding and Data Caching}
\label{sec:revisit}

In conventional layer freezing methods, to ensure proper training of the unfrozen portions of the network, we need to provide training data to these unfrozen layers. Specifically, the original data is passed through the frozen layers via forward propagation to generate intermediate outputs, which then serve as training data for the unfrozen layers. This process turns the frozen portion of the network into a data generator. 
In the hypothetical solution (i.e., caching) proposed in \cite{b8,b12}, all the intermediate outputs of the frozen layer (generated over the entire original training dataset) are cached as a new dataset and stored on disk storage.
Then, the unfrozen layers can directly load the training data from the new dataset and skip the frozen layers. However, there are several \textbf{critical missing pieces} in the hypothetical solution.

\textbf{Data augmentation issue.}
First, the original caching method does not address the data augmentation issue.
The caching method stores intermediate data (i.e., activations/feature maps) based on the original (non-augmented) training data, and also no augmentation is applied to the loaded cached data (feature map) to train the unfrozen layer. As a result, a severe accuracy degradation can be observed in the model accuracy \cite{b8}.
Unfortunately, this is not a trivial problem to solve. 
We conducted comprehensive experiments and found that by directly applying existing augmentation methods to the loaded cached data, including random cropping, random flipping, rotation, and color jitter \cite{b22,b23}, as well as the spatial transformations and noise perturbations \cite{b2,b3}, none of those augmentations can effectively prevent accuracy degradation.
Therefore, we design a new augmentation method, \textit{similarity-aware channel augmentation}, to effectively augment loaded data and preserve accuracy for CNNs (Sec. \ref{sec:similarity-aware aug}) and transformer-based models (Sec. \ref{sec:aug_transformer}).

\textbf{Substantial storage requirements.}
The original caching method directly stores all the intermediate data.
Since intermediate data typically have a larger size than the original input data due to the increased dimensionality and higher representation precision (e.g., from integers to floating-point numbers), the newly stored dataset will have a substantially larger size than the original dataset, introducing significant even unacceptable storage overhead.

It seems that the lossy compression method is a promising solution for this issue, but there is no prior work that explores the actual potential of it.
In this work, we explore the impact of lossy compression ratio and compression chuck size on the model accuracy, memory saving, and training acceleration (Sec. \ref{sec:Experiment}). 
Then, we propose a progressive compression method, which reduces storage and computational overhead while maintaining model performance (Sec. \ref{sec:progressive_compress}).
Moreover, we propose a coarse-grained chuck shuffling method to efficiently shuffle the cached training data while achieving a good balance between decompression speed and model accuracy (Sec. \ref{sec:Chuck_Shuffling}).
Figure \ref{fig:pipeline} shows the overview of our proposed methods.

\begin{figure*}[!t]
\centering
        \includegraphics[width=0.85\textwidth]{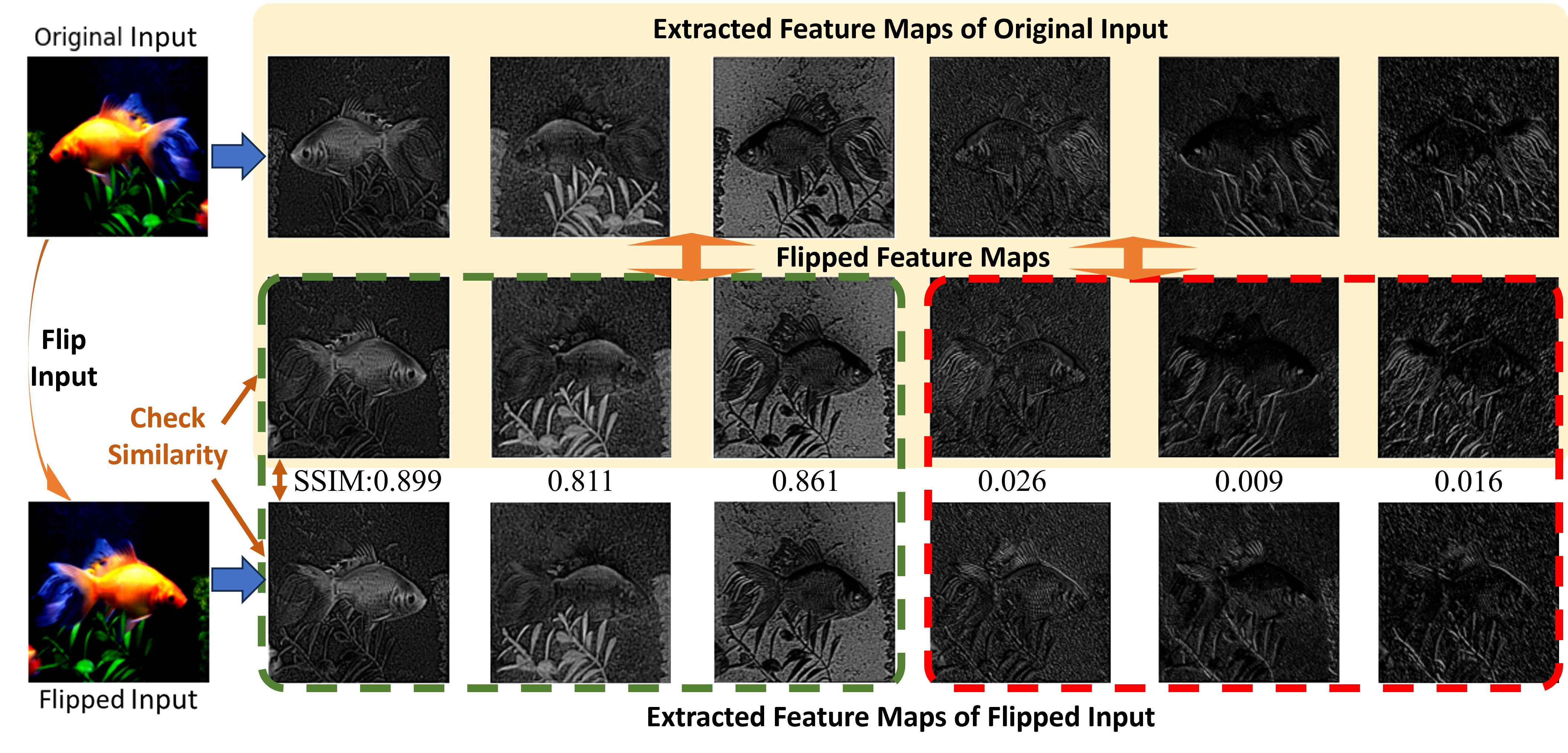} 
        \caption{The sensitivity of different convolution kernels to spatial transformations. The first row shows the selected activation channels for the original input; the second row presents the flipped feature maps of original input; the third row displays the feature maps of flipped input.}
    \label{fig:combined}
\end{figure*}
\begin{figure*}[!t]
\centering
\includegraphics[width=0.9\textwidth]{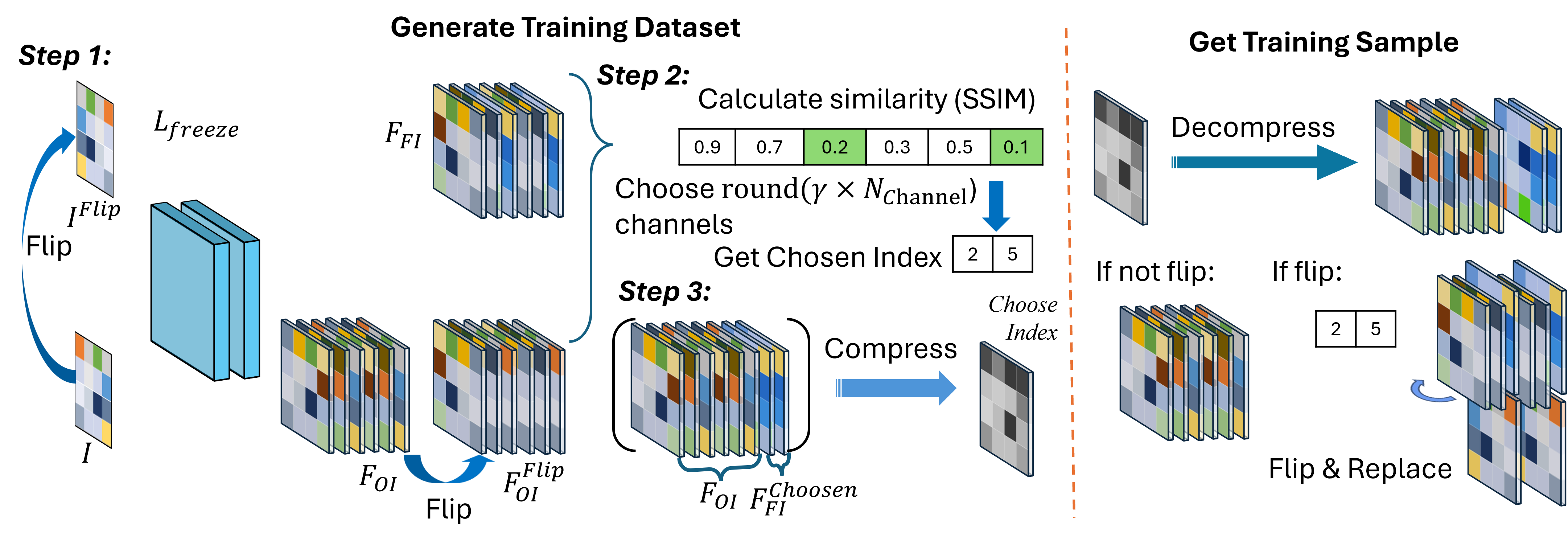}
\caption{The overview of the Similarity-Aware Channel Augmentation.}
\label{fig:flip_aug}
\end{figure*}

\subsection{Similarity-Aware Channel Augmentation}
\label{sec:similarity-aware aug}

Inspired by \cite{b4}, the information processed by each convolutional kernel is distinct. Our experiments also reveal that convolutional kernels exhibit varying degrees of sensitivity to spatial transformations. 

For kernels insensitive to spatial transformations (e.g., left three feature maps in Figure \ref{fig:combined},flipping the input directly results in a corresponding flip in activation, making the transformed and original feature maps nearly same) the  spatial transformations can be directly applied. For kernels that are sensitive to such transformations (as shown in Figure \ref{fig:combined} the right three feature maps, the flip operation will result in different feature maps, as seen by the clearly difference between the second row and the third row), alternative strategies are required.

Building on this observation, we propose \textbf{Similarity-Aware Channel Augmentation}, a novel method that addresses the issue by storing transformation information to enhance activations. Specifically, as illustrated in Figure \ref{fig:flip_aug}, for sensitive kernels, we cache their activations and use the cached values to replace corresponding channels during flip-based augmentation, forming a new input tensor.


The process begins by identifying the channels to store based on the similarity between the flipped feature maps of the original input and those of the flipped input. First, flip the original input tensor $ I $ to obtain $ I^{{Flip}} $. Pass both $ I $ and $I^{{Flip}}$ through the frozen layer $ L_{freeze} $ to obtain feature maps $ F_{OI} $ and $ F_{FI} $. Then, flip $ F_{OI} $ to generate $ F_{OI}^{Flip} $ and calculate the similarity \textbf{SSIM} between $  F_{FI} $ and $ F_{OI}^{Flip} $ for each channel. Finally, select the $ x $ most dissimilar channels (with the lowest SSIM) denotes as $F_{FI}^{Choosen}$, for storage, along with original feature maps $F_{OI}$. The value of $ x $ is determined by the hyperparameter $ \gamma $ as $ x = \text{round}(\gamma \cdot c) $, with $ c $ denotes the total number of channels.

During Post-Discarding stage, the stored activations are used to replace the sensitivity channels during flip-based augmentation. First decompress the activations. When flip augmentation is applied, flip the original tensor $T$ to obtain $T_{Flip}$ and replace the specified channels in $T_{Flip}$ with $T^{Flip}_{Choosen}$ to form a new tensor. Additional augmentation techniques, such as random cropping, adding Gaussian noise, or Gaussian blur, can be applied to the augmented tensor.

This method mitigates the impact of spatial transformations on sensitive kernels by storing and restoring key activations. It identifies and retains the most sensitive channels, replacing them during flip-based augmentation to prevent distortion while allowing additional transformations, thereby enhancing the model’s generalization ability.

\subsection{Augmentation for Transformer Models }
\label{sec:aug_transformer}
\vspace{-0.5em}
\begin{algorithm}[b]
\caption{The framework of generate the augmentated token for one training sample.}
\label{alg:cosine_similarity_ranking}
\textbf{Input:} Original token set $T_{\text{ori}}$, augmented token set $T_{\text{aug}}$.\\
\textbf{Output:} Token $t_{\text{aug}}$ with the lowest cosine similarity.

\begin{algorithmic}[1]
\State Initialize $min\_cosine \gets \infty$, $best\_pair \gets \text{None}$,

\ForAll{$i \in N_{token} $ }
    \ForAll{$j \in N_{token} $}
        \State $t_{\text{ori}} = T_{\text{ori}^{i}}$ and $t_{\text{aug}} \in T_{\text{aug}}$
        \State Compute cosine similarity $s(t_{\text{ori}}, t_{\text{aug}})$
        \Comment{$s(t_{\text{ori}}, t_{\text{aug}}) = \frac{t_{\text{ori}} \cdot t_{\text{aug}}}{\|t_{\text{ori}}\| \|t_{\text{aug}}\|}$}

        \If{$s(t_{\text{ori}}, t_{\text{aug}}) < min\_cosine$}
            \State $min\_cosine \gets s(t_{\text{ori}}, t_{\text{aug}})$
            \State $best\_pair \gets (i, j,s(t_{\text{ori}}, t_{\text{aug}}) )$
        \EndIf
    \EndFor
    \State Append $best\_pair$ to $pairs$
\EndFor
\State Sort $pairs$ by $min\_cosine$ in ascending order
\State Select the lowest $\alpha$ percentage of pairs from $pairs$

\State Extract the corresponding tokens of $T_{\text{aug}}^{Selected}$ from the selected pairs
\State \Return $T_{\text{aug}}^{Selected}$ and $pairs$
\end{algorithmic}
\end{algorithm}
For transformer-based models, we applied a similar strategy of storage and replacement enhancement. However, since the activations in transformer-based models are composed of tokens and lack the spatial structure inherent in CNN-based activations, techniques such as flip and clip cannot be directly applied. Instead, we adopted a token replacement approach.

In this approach, we first identify the differences between tokens generated from augmented and those from original images. We then select the most divergent tokens for storage and replace them during data augmentation, thereby introducing augmentation information into the tokens.

Given that positional embeddings cause the token order to change under spatial transformations, we calculate the cosine similarity between each original token $T_{\text{ori}}^i$ and all augmented tokens $T_{\text{aug}}^{1 \sim \text{Token\_num}}$. For each original token, the augmented token $T_{\text{aug}}^j$ with the highest similarity is matched. Next, we select the bottom $\alpha\%$ of tokens (those with the lowest similarity) for storage. During augmentation, these stored tokens are used to replace the corresponding tokens in the augmented sequence. The detailed process is described in Algorithm~\ref{alg:cosine_similarity_ranking}.

\subsection{Progressive Compression}
\label{sec:progressive_compress}
As mentioned in Sec. \ref{sec:revisit}, activations often require significantly more storage space than the original data.
As models grow deeper, this storage demand becomes even more pronounced, making activation compression essential to reduce overhead.
Fortunately, deeper layers naturally present opportunities for compression. 
As data propagates through the network, many low-level details irrelevant to the final task are gradually ``filtered out'', leading to more concentrated meaningful information in feature maps. The increased redundancy and sparsity create favorable conditions for compression algorithms to achieve a higher compression ratio. 

Based on this insight, we propose \textbf{Progressive Compression}, a technique that adaptively reduces activation sizes to effectively balance storage overhead and model performance. Our approach utilizes the ZFP algorithm, which supports efficient lossy compression while preserving numerical precision. To optimize space efficiency, we adjust the compression ratio based on layer depth. Specifically, we introduce a tolerance parameter, $\tau$, to determine the appropriate compression ratio for each layer. For deeper layers, it achieves a higher compression ratio without performance degradation by setting $\tau$.

To set the $\tau$, we investigate the compressibility of activations, by setting an acceptable tolerance level for each layer and evaluating the maximum compression ratio achievable under that tolerance, as detailed in Section \ref{sec:cmp_ratio}.  Our findings show that deeper layers produce more compressible activations under the same tolerance, enabling greater storage savings. Thus, we can make the deeper layer with a higher compression ratio and simply set the same tolerance for all layers. This approach allows the model to retain its original accuracy during training with decompressed feature maps.

\subsection{Coarse-grained Chuck Shuffling}
\label{sec:Chuck_Shuffling}
Since intermediate data are cached as a whole, they must be decompressed together, limiting the flexibility of shuffling training data, which can also degrade model performance. 
A straightforward solution is to compress each sample individually and decompress them on demand, but this introduces additional time overhead due to frequent decompression operations. 
To balance model performance and efficiency, we propose to adopt coarse-grained chuck shuffling, where multiple samples are compressed together. However, excessively large blocks may degrade training performance, while overly fine blocks increase decompression overhead. 
To strike a balance, we compress two samples per block, with the selection criteria detailed in Appendix ~\ref{app:speed}..
\vspace{-0.5em}

\section{Experiment}
\label{sec:Experiment}

\textbf{Experiment setup.}
To validate our methods, we conducted experiments using CNN-based model ResNet32, ResNet50 \cite{resnet}, and transformer-based model DeiT~\cite{b16} in both \textit{train-from-scratch} and \textit{fine-tuning} scenarios, and all of our experiments are conduct on a single NVIDIA A100 GPU.

\textit{For fine-tuning},
\textbf{ResNet32} was pre-trained on CIFAR-100 \cite{b30} and fine-tuned on the CIFAR-10 dataset for 80 epochs. The initial learning rate was set to 0.01, and all other settings were the same as those used in the train-from-scratch scenario. \textbf{ResNet50} was pre-trained on ImageNet \cite{imagenet} was fine-tuned on the CIFAR-100 dataset for 50 epochs. The settings were identical to those used for ResNet32 fine-tuning. The \textbf{DeiT(small)} was pre-trained on ImageNet and fine-tuned on the CIFAR-100 dataset for 100 epochs. The initial learning rate was set to $6 \times 10^{-5}$ and adjusted using cosine decay. All other settings were consistent with those used in the ResNet50 fine-tuning scenario.

\textit{For training from scratch,}
\textbf{ResNet32} was trained on CIFAR-100 dataset for a total of 160 epochs. The initial learning rate was set to 0.1 and adjusted using cosine decay. The optimizer was SGD with momentum set to 0.9. The batch size was 64. \textbf{ResNet50} was trained on Tiny-ImageNet \cite{tinyimagenet} dataset for 100 epochs. All other settings, including learning rate, optimizer, and batch size, were identical to those used for ResNet32.

\vspace{-0.5em}
\subsection{Definitions of Freezing Strategies and Metric}
This section give an explanation of model freezing strategies in Table \ref{tab:performance_finet} and Table \ref{tab:performance_train} and the metric we used to show the performance. 


\textbf{Normal} refers to standard training or fine-tuning, where the entire model is trained without any parts being frozen. \textbf{Ours w/ Discard} discards the frozen part and augments activations using standard techniques, such as direct feature map flipping and random cropping. \textbf{Ours w/ Discard \& Aug.} applies our proposed \textit{Similarity-Aware Channel Augmentation}. Measurement details for memory usage, FLOPs, and speed are provided in Appendix~\ref{app:mem_measure}. Accuracy is reported as the average over three runs with different random seeds, along with standard deviation.
Note that activations without any data augmentation significantly degrade performance and are therefore excluded from the table. For this proof-of-concept, we adopt the same freezing strategy as FreezeOut \cite{b15}, though our method is also compatible with other freezing and compression approaches.

\vspace{-0.5em}
\subsection{Results on Fine-tuning}

\begin{table*}[t]
\centering
\caption{Comparison of accuracy, average memory, minimum memory, and training FLOPs across different freezing methods and datasets in finetuning a model.}
\scalebox{0.75}{
\begin{tabular}{@{}>{\centering\arraybackslash}m{1.2cm}>{\centering\arraybackslash}m{1.25cm}>{\centering\arraybackslash}m{3.5cm}>{\centering\arraybackslash}m{1.5cm}>
{\centering\arraybackslash}m{1.3cm}>
{\centering\arraybackslash}m{1.3cm}>
{\centering\arraybackslash}m{1.4cm}@{}}
\toprule
\textbf{Model} & \textbf{Dataset} & \textbf{Method} & \textbf{Accuracy (\%)} & \textbf{Avg. Memory (MB)}& \textbf{Min. Memory (MB)} & \textbf{FLOPs ($\times10^{15}$)} \\ 

\midrule
\multirow{4}{*}{ResNet32} 
    & \multirow{4}{*}{\shortstack{CIFAR100 \\ $\downarrow$ \\ CIFAR10}}
      & Normal                  & 93.07 & 630  & 630   & 6.48  \\ 
    & & FreezeOut \cite{b15}     & 92.91 & 409 & 364   & 4.05  \\ 
    & & Ours w/ Discard         & 92.85$\pm$0.26 & 188 & 99   & 2.92  \\ 
    & & Ours w/ Discard \& Aug. & 93.06$\pm$0.21 & 188 & 99   & 2.92  \\ 
\midrule
\multirow{5}{*}{ResNet50} 
    & \multirow{5}{*}{\shortstack{ImageNet \\ $\downarrow$ \\ CIFAR100}} & Normal & 82.60 & 5002 & 5002  & 19.09 \\ 
    & & FreezeOut \cite{b15}              & 82.35 & 3372 & 2701  & 12.85 \\ 
    & & SmartFRZ \cite{b9}               & 81.95 & 2241 & 1834   & 9.88 \\ 
    & & Ours w/ Discard               & 82.26$\pm$0.19 & 1743 & 402   & 9.73  \\ 
    & & Ours w/ Discard \& Aug.  & 82.43$\pm$0.18 & 1743 & 402  & 9.73 \\ 

\midrule
\multirow{5}{*}{DeiT} 
    &\multirow{5}{*}{\shortstack{ImageNet \\ $\downarrow$ \\ CIFAR100}} & Normal & 89.00 & 15095 & 15095  & 125.10\\ 
    & & FreezeOut \cite{b15}          & 88.62 & 11190 & 9088  & 84.25 \\ 
    & & SmartFRZ \cite{b9}           & 88.52 & 9811    & 7849     & 72.93 \\ 
    & & Ours w/ Discard          & 88.37$\pm$0.56 & 7723 & 3754  & 63.84 \\ 
    & & Ours w/ Discard \& Aug.  & 88.43$\pm$0.38 & 7723  & 3754 & 63.84 \\ 
\bottomrule
\end{tabular}
}
\label{tab:performance_finet}
\end{table*}

As shown in Table \ref{tab:performance_finet}, our approach significantly reduces FLOPs and memory usage without degrading model performance.
For \textbf{ResNet32}, our method achieved a slight improvement in Top-1 accuracy while reducing FLOPs by 28\% and memory usage by 54.1\% compared to the baseline.
For \textbf{ResNet50}, our method outperformed both the baseline and SmartFrz, reduce 24.4\%  FLOPs and  48.4\%  in memory usage compared to the baseline, while also reducing memory consumption by 22.3\% compared to SmartFrz.
For \textbf{DeiT(small)}, our method maintained comparable performance while reducing computation costs by 24.3\% and memory usage by 31.3\%. Additionally, it achieved similar accuracy with 21.3\% less memory usage and 12.5\% lower computation costs.
Experimental results demonstrate that our method significantly reduces the computational cost of fine-tuning while maintaining model performance.

\vspace{-0.5em}
\subsection{Results on Training from Scratch}

\begin{table*}[h]
\centering
\caption{Comparison of accuracy, average memory, minimum memory, and training FLOPs across different freezing methods and datasets in training a model from scratch.}
\scalebox{0.75}{
\begin{tabular}{@{}>{\centering\arraybackslash}m{1.2cm}>{\centering\arraybackslash}m{1.25cm}>{\centering\arraybackslash}m{3.5cm}>{\centering\arraybackslash}m{1.8cm}>
{\centering\arraybackslash}m{1.3cm}>
{\centering\arraybackslash}m{1.3cm}>
{\centering\arraybackslash}m{1.4cm}@{}}
\toprule
\textbf{Model} & \textbf{Dataset} & \textbf{Method}  & \textbf{Accuracy (\%)} & \textbf{Avg. Memory (MB)} & \textbf{Min. Memory (MB)} & \textbf{FLOPs ($\times10^{15}$)} \\ 
\midrule
\multirow{4}{*}{ResNet32}
    & \multirow{4}{*}{CIFAR-100}& Normal   & 74.62 & 630 & 630     & 12.97 \\ 
    & & FreezeOut \cite{b15}                & 73.76 & 486 & 453    & 10.47 \\ 
    & & Ours w/ Discard              & 73.14$\pm$0.08 & 344 & 278    & 9.30  \\ 
    & & Ours w/ Discard \& Aug.  & 73.61$\pm$0.12 & 344 & 278    & 9.30  \\ 

\midrule
\multirow{4}{*}{ResNet50} 
    &\multirow{4}{*}{\shortstack{Tiny-\\ImageNet}} & Normal            & 64.89 & 19465 & 19465    & 305.47 \\ 
    & & FreezeOut \cite{b15}         & 64.08 & 13985 & 10360   & 219.63\\ 
    & & Ours w/ Discard       & 63.18$\pm$0.34 & 8508 & 1259   & 176.69 \\ 
    & & Ours w/ Discard \& Aug. & 63.42$\pm$0.22 & 8508 & 1259    & 176.69 \\ 

\bottomrule
\end{tabular}
}
\vspace{-1.5em}
\label{tab:performance_train}
\end{table*}

As shown in Table \ref{tab:performance_train}, our method also demonstrates significant memory and FLOP savings compared to the baseline.  
For \textbf{ResNet32},  resulting in an 11.2\% reduction in FLOPs and a 20.3\% reduction in memory usage, with only a 0.15\% drop in Top-1 accuracy.  
For \textbf{ResNet50}, our method reduced FLOPs by 19.6\%  and a 39\% reduction in memory usage, with a minimal performance drop of 0.66\% in Top-1 accuracy.

\subsection{Training Speed}
\begin{wrapfigure}{r}{0.53\textwidth}

    \centering
    \includegraphics[width=\linewidth]{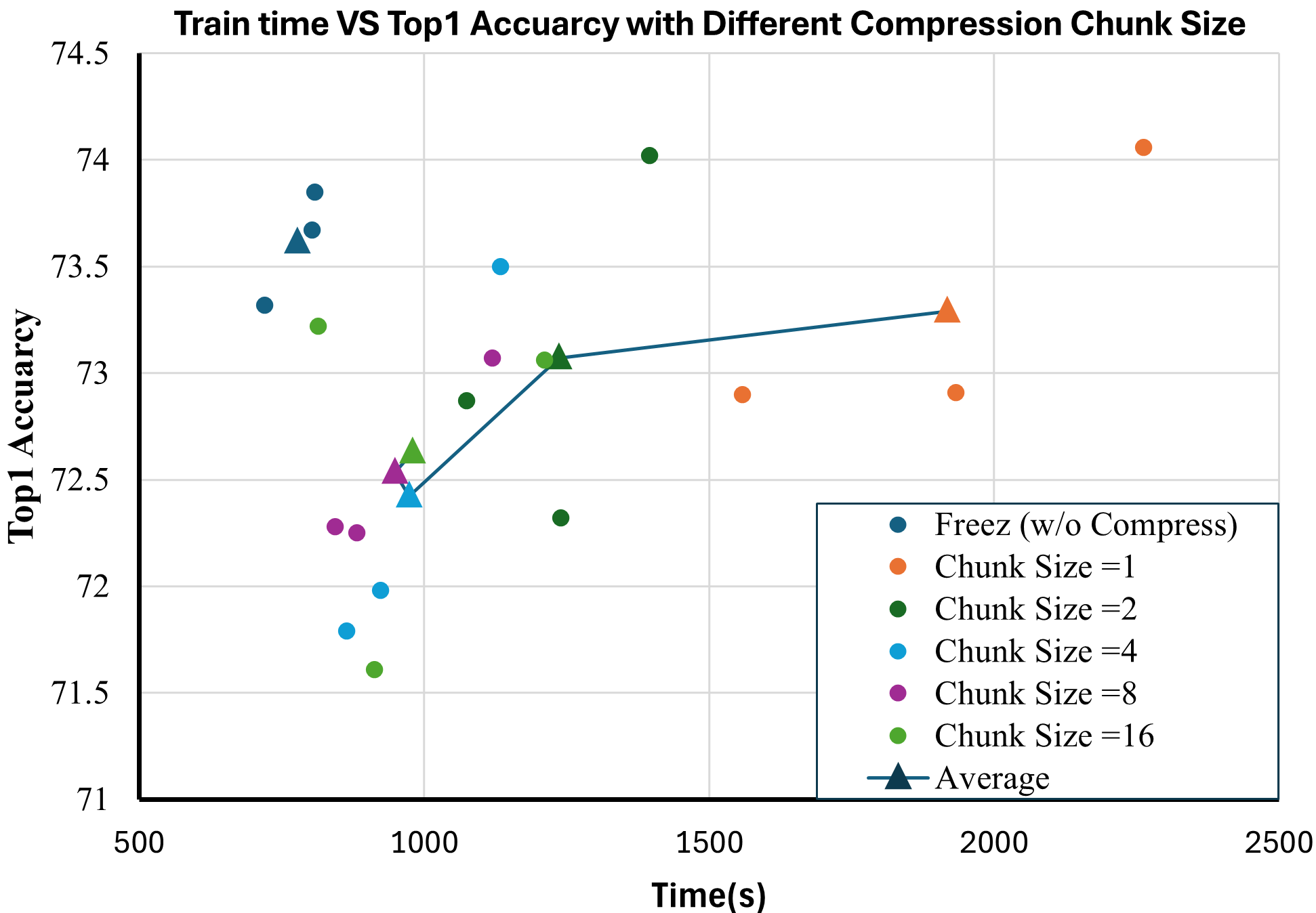}
    \caption{Experiment with different compression chunk size on ResNet 32, each chunk size test with 3 frozen strategy frozen, frozen 1st block at 30 epochs, 2nd block at 60 epochs, and 1st \& 2nd block at 30 and 60 epochs separately.}
    \label{fig:acc_vs_time}

\end{wrapfigure}
The introduction of compression and decompression during training adds extra time overhead, primarily due to decompression operations being limited to CPU. This increases data loading and transfer times, leading to longer training durations.
To reduce this overhead, we employed \textbf{coarse-grained compression and decompression}, where multiple samples are compressed and decompressed in a single operation, so that it could accelrate the training process. 
As shown in Figure \ref{fig:acc_vs_time}, the individual compress cost the most time, and when the compression chunk size increases to 2, the average time decreased from 1917 seconds to 1236 seconds. And when the chunk size increase to 4,8,16 the time cost going shorter.

\subsection{Ablation Study}
\label{stu:cmp_size}

\textbf{Compression analysis.} Our experiments shows that feature maps at deeper layers and later epochs could achieve higher compression rates under the same tolerance. The study is detailed in \ref{sec:cmp_ratio}.

\textbf{Training speed analysis.} Although the \textbf{coarse-grained compression and decompression} significantly reduces overhead but also lowers the randomness of the data, which can slightly degrade model performance.
Our experiments show that reducing the frequency of compression and decompression operations can achieve training times comparable to baseline methods. This indicates that the acceleration gained from our approach effectively offsets the time overhead introduced by compression.
The analysis of tradeoff between time and accuracy  is detailed in Appendix ~\ref{app:speed}.
\vspace{-1em}
\subsection{Discussion and Limitations}
\label{limitation}
In this paper, we have revealed that the caching technique for layer freezing can significantly save memory costs and training FLOPs while maintaining accuracy, which is particularly valuable for resource and efficiency considerations.
However, it still shows a slightly compromised training speed due to the limited computing speed of current data compression techniques. In our experiments, we use the ZFP compression method, and the CPU is used to compute the compression/decompression process.
Nevertheless, by incorporating a dedicated system optimization for the compression method (e.g., using GPU) or with the advancement of data compression techniques in the future, the potential of the caching technique for layer freezing will be further unleashed.
And our comprehensive study provides insightful inspiration for future research.

\vspace{-1em}
\section{Conclusion}

In this paper, we comprehensively explore the overlooked issues in caching techniques for layer freezing. We propose a similarity-aware channel augmentation method to effectively improve the model accuracy. Moreover, we explore the compatibility and potential of incorporating lossy compression in caching.
Our findings provide practical insights for balancing speed and performance when using freezing and compression techniques. These strategies can be tailored to specific tasks, enabling a more efficient and effective training process.


\appendix
\renewcommand{\thefigure}{\Alph{section}.\arabic{figure}}  
\renewcommand{\thetable}{\Alph{section}.\arabic{table}}    
\counterwithin{figure}{section}  
\counterwithin{table}{section}

\section{Appendix / supplemental material}

\subsection{Memory, Flops, and Speed Measurement}
\label{app:mem_measure}
The memory values shown in Table \ref{tab:performance_finet} and Table \ref{tab:performance_train} include the \textbf{Average Memory} which show the mean value of memory usage throughout the entire training process, and the \textbf{Minimum Memory} indicates the memory costs of the training after all frozen layers have been finalized.

The memory is computed by the memory usage at each freezing status times the number of epochs of that status. 
FLOPs are calculated in a similar manner by measuring the per-sample FLOPs at each stage using ptflops \cite{b26}, multiplying by the total number of samples and the number of epochs for each stage, and summing the results. Additional FLOPs incurred by activation generation and decompression in discard methods are also included. The computational cost, measured in FLOPs, and the memory usage under different freezing states are summarized in Table~\ref{tab:memory_freeze_drop} and Table~\ref{tab:flops_freezing}, respectively, the position in the table represent the $n_{th}$ corresponding residual block or attention block. The saved FLOPs represent the reduction in computation from the forward pass due to discarding at specific positions, rather than from standard layer freezing. The remaining FLOPs indicate the computational cost required after the discarding has taken place. The Flops overhead of our Similarity Aware Channel Augmentation is just a forward process of frozen layers for whole dataset, which is a very small part to the whole training or finetuning process.

\begin{table}[htbp]
\centering
\caption{Memory usage at different freeze (discard) positions. The unit of memory is MB.}
\renewcommand{\arraystretch}{1.0}
\begin{tabular}{@{}>{\centering\arraybackslash}m{1.2cm}>{\centering\arraybackslash}m{1.8cm}>{\centering\arraybackslash}m{2.5cm}>{\centering\arraybackslash}m{2.5cm}>
{\centering\arraybackslash}m{2.2cm}@{}}
\hline
\textbf{Model} & \textbf{Total Memory} & \textbf{Freeze or Discard Position}  & \textbf{Just Freezing Memory } & \textbf{With Discard Memory} \\ 
\midrule
\multirow{2}{*}{ResNet32} 
    &\multirow{2}{*}{630} & 5  & 454 & 278 \\ 
    & & 10  &   364 & 99  \\ 

\midrule
\multirow{3}{*}{ResNet50} & \multirow{3}{*}{5002}
    & 3   & 3945 & 2889 \\ 
    & & 7                              & 3228 & 1456 \\

    & & 13                              & 2701 & 402  \\ 
\midrule
DeiT & 15096 & 9  & 9088 & 3754 \\ 

\bottomrule
\end{tabular}
\label{tab:memory_freeze_drop}
\end{table}

\begin{table}[htbp]
\centering
\caption{FLOPs distribution under different freeze (discard) positions. The value in the table is the FLOPs of a single data point, the unit of FLOPs in this table is GFLOPS.}
\renewcommand{\arraystretch}{1.0} 
\setlength{\tabcolsep}{4pt} 
\begin{tabular}{@{}>{\centering\arraybackslash}m{1.2cm} >{\centering\arraybackslash}m{2.5cm} >{\centering\arraybackslash}m{2.5cm} >{\centering\arraybackslash}m{2.5cm}>{\centering\arraybackslash}m{1.0cm}>{\centering\arraybackslash}m{1.0cm}@{}}
\toprule
\textbf{Model} & \textbf{Input Size}  & \textbf{Training FLOPs w/o Freeze} & \textbf{Freeze / Discard Position} & \textbf{Saved FLOPs} & \textbf{Remain FLOPs}   \\
\midrule

\multirow[c]{2}{*}{ResNet32} 
    & \multirow[c]{2}{=}{\centering $1 \times 3 \times 32 \times 32$} & \multirow[c]{2}{=}{\centering 1.66} & 5  & 0.19 & 1.08\\ 
    &  &  & 10 & 0.36 & 0.53 \\
\midrule

\multirow[c]{3}{*}{ResNet50} 
    & \multirow[c]{3}{=}{\centering $1 \times 3 \times 32 \times 32$} & \multirow[c]{3}{=}{\centering 7.82} & 3  & 0.43 & 6.49\\ 
    &  &  & 7  & 1.12 & 4.46\\ 
    &  &  & 13 & 2.08 & 1.59\\ 
\midrule

DeiT & $1 \times 3 \times 224 \times 224$ & 25.62 & 9 & 6.43 & 6.32\\ 
\bottomrule

\end{tabular}
\label{tab:flops_freezing}
\end{table}

The training speed is measured by the total time cost of the training process, which includes both the standard training duration and the overhead caused by data compression and decompression. The experimental results are shown in Figure \ref{fig:acc_vs_time}.

\subsection{Result of compression}
\label{sec:cmp_ratio}

\subsubsection{\textbf{Higher compression rates for later epoch}}
\begin{figure}[h]
\centering
\includegraphics[width=0.99\textwidth]{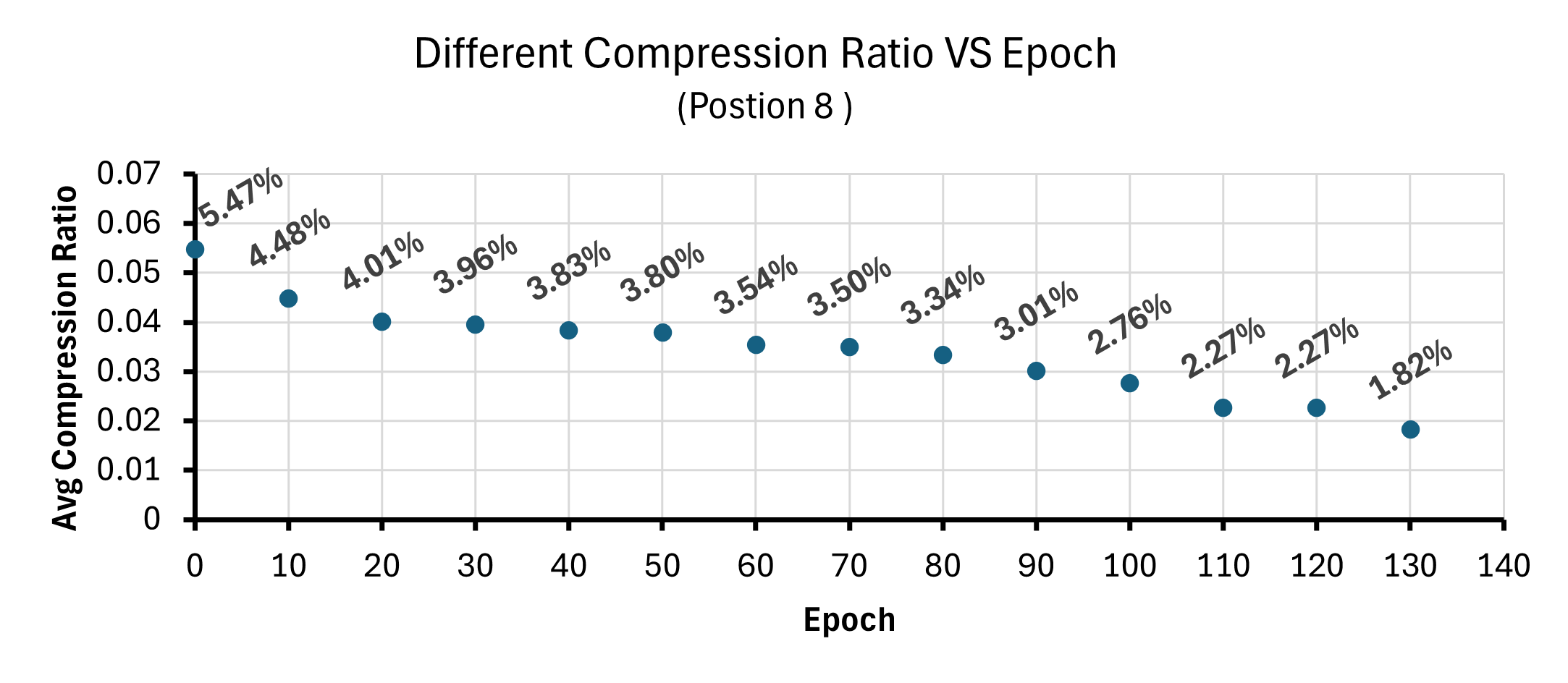}
\caption{Feature maps compression rates at same position for different training epoch.}
\label{fig:Cmp_time}
\end{figure}
We also found that within the same layer, the compression rate increases as training progresses. For example, as shown in Figure \ref{fig:Cmp_time}, the output of the 8th residual block achieves a higher compression rate in later training stages.

\subsubsection{\textbf{Higher compression rates for deeper layers}}
\begin{figure}[h]
\centering
\includegraphics[width=0.99\textwidth]{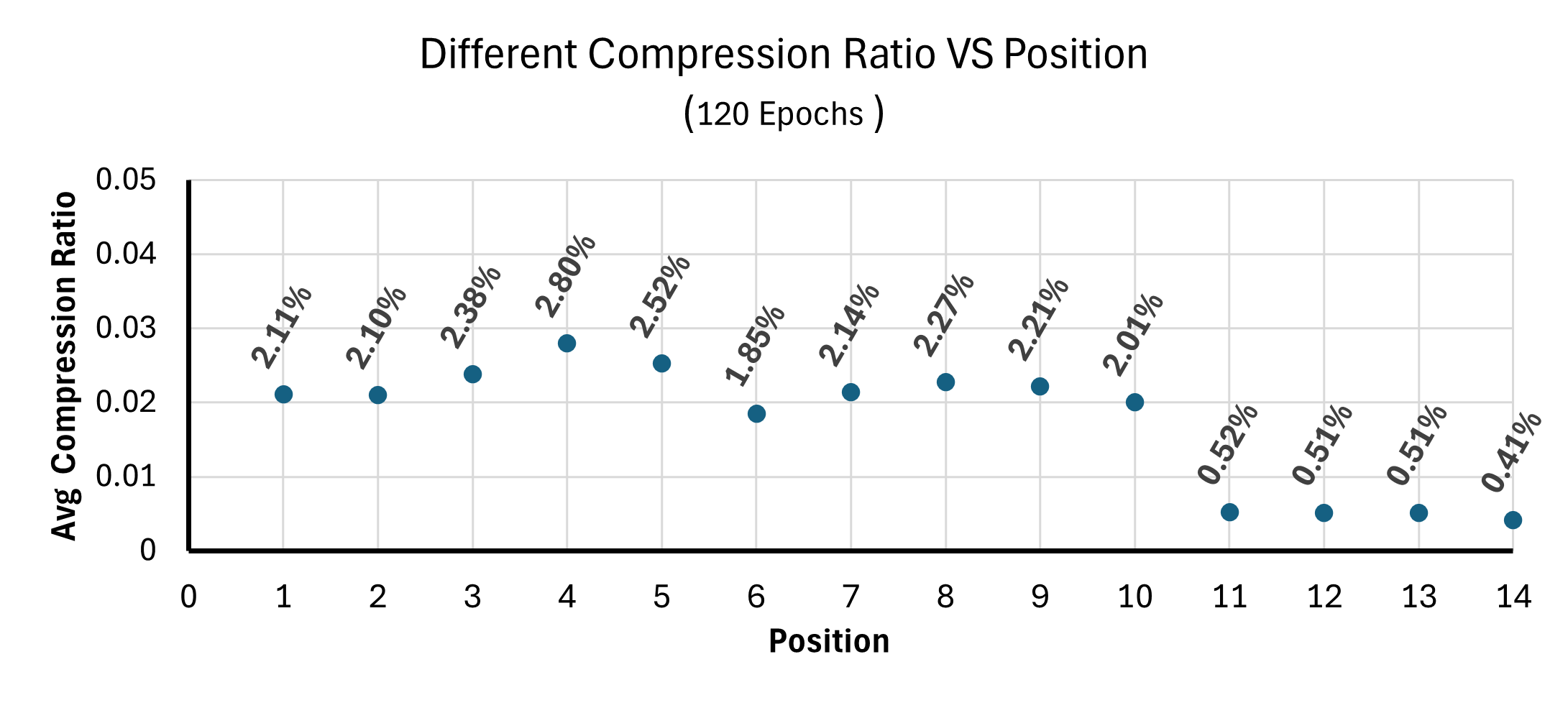}
\caption{Feature maps compression rates at same training epoch for different model position.}
\label{fig:Cmp_pos}
\end{figure}
As shown in Figure \ref{fig:Cmp_pos}, when the tolerance fixed and the model is trained for 100 epochs, activations from deeper layers achieve significantly higher compression rates. 

\subsubsection{\textbf{Higher compression rates for bigger chunk size}}
\begin{table}[h]
\centering
\caption{Compression rates and compression and decompression time (in seconds) cost under different compression chunk sizes. The experiment is conducting with ResNet32 trained on CIFAR-100 dataset. Note that compression only needs to be conducted once when generating the dataset.}
\renewcommand{\arraystretch}{1.0} 
\setlength{\tabcolsep}{4pt} 
\begin{tabular}{@{}>{\centering\arraybackslash}m{1cm} >{\centering\arraybackslash}m{2.5cm} >{\centering\arraybackslash}m{1cm} >{\centering\arraybackslash}m{1cm} >{\centering\arraybackslash}m{1cm} >{\centering\arraybackslash}m{1cm} >{\centering\arraybackslash}m{1cm}@{}}

\toprule
\multirow[c]{2}{*}{\textbf{Position}} & \multirow[c]{2}{*}{\textbf{Metric}} & \multicolumn{5}{c}{\textbf{Compress/Decompress Chunk Size}} \\ 
\cmidrule{3-7}
& & \textbf{1} & \textbf{2} & \textbf{4} & \textbf{8} & \textbf{16}\\  
\midrule

\multirow[c]{3}{*}{5} 
& Compress Rates  & 0.197 & 0.117 & 0.064 & 0.058 & 0.058 \\
& Compress time        & 94.93 & 53.62 & 31.41 & 24.38 & 21.85 \\ 
& Decomp. Time & 0.0005 & 0.0006& 0.0007 & 0.0013 & 0.0028 \\
\midrule

\multirow[c]{2}{*}{10} 
& Compress Rates  & 0.172 & 0.105 & 0.055 & 0.055 & 0.055 \\ 
& Compress Time        & 100.21 & 55.62 & 31.07 & 22.00 & 17.44 \\ 
& Decomp. Time & 0.0002 & 0.0003 & 0.0003 & 0.0006& 0.0012 \\
\bottomrule
\end{tabular}
\label{tab:cmp_ratio}
\end{table}
Table \ref{tab:cmp_ratio} summarizes the effect of different compression/decompress chunk sizes (the number of activations compressed together in a single operation). Under the same tolerance, larger chunk sizes lead to better compression rates but higher compression/decompression time. 

With the compress chunk size of two, the feature maps will compress to the 1/10 of original size while keep the performance, that means in some freezing strategy, this will be smaller than original dataset, and will reduce the disk memory cost. For example, for the DeiT trained on $224\times224\times3$ image the token size will be $384\times198\times4$ which is only $2.09\times$ of original data size, after compress this will become $0.23\times$ of original dataset.



\subsection{Trade-off between Speed and Performance}
\label{app:speed}

 As shown in Figure\ref{fig:acc_vs_time}, the individual compress have the best performance, and when the compression chunk size increases to 2, there is a little performance degrade of about 0.2\%. And when the chunk size increase to 4,8,16 the accuracy degrade from 73.3 to 72.5. Thus, in this situation, a chunk size of 2 is a suitable choice, as it does not cause significant performance degradation and avoids substantial time overhead. In practical applications, the compression chunk size should be selected based on the specific trade-off between training speed and accuracy. This trade-off between compression efficiency and model accuracy is conceptually similar to what has been observed in feature-space augmentation methods such as FMix~\cite{b28} and SuperMix~\cite{b29}.

 Currently, the performance of our method is bottlenecked by the speed of the compression algorithm itself. Future, integration with faster compression techniques could further enhance training efficiency, as improvements in compression speed would directly compound the benefits of our framework.


\end{document}